\documentclass[letterpaper,conference]{IEEEtran}

\IEEEoverridecommandlockouts
\usepackage{amsmath,amssymb,amsthm,bm,booktabs,graphicx}
\usepackage{microtype}
\usepackage{xspace}
\usepackage{hyperref}
\hypersetup{
  colorlinks,
  linkcolor={red!50!black},
  citecolor={blue!50!black},
  urlcolor={blue!80!black},
}
\usepackage{outlines}
\usepackage{xcolor}
\usepackage{paralist}
\usepackage[subtle]{savetrees}
\usepackage{titlesec}
\titlespacing{\subsection}{1.5pt}{\parskip}{\parskip}
\newcommand{\SE}{\mathrm{SE}(2)}
\newcommand{\name}{RPI\xspace}
\newtheorem{proposition}{Proposition}
\title{Radio-to-Pose Information for Gauge-Drift Correction in Cooperative Localization}
\author{
\IEEEauthorblockN{
Aditya Iyer,
Andy Ye,
Akshay Narayan,
Nora Ayanian
}
\IEEEauthorblockA{
\{aditya\_iyer, akshayn, nora\_ayanian\}@brown.edu\\
andy132@cs.washington.edu
}
}
\usepackage[
  letterpaper,
  left=0.75in,
  right=0.75in,
  top=0.75in,
  bottom=0.80in,
  headheight=0pt,
  headsep=0pt
]{geometry}

\microtypesetup{protrusion=false}
\begin{document}
\maketitle
% \raggedbottom
\thispagestyle{empty}
\pagestyle{empty}

\begin{abstract}
Cooperative localization uses odometry and inter-robot measurements to
estimate the motion of a robot team. 
These measurements describe only relative geometry, but the 
team's estimated position and orientation can drift together over time, \emph{i.e.}, \textit{gauge drift}. 
In this work, we study correction of this gauge drift using a radio
attenuation map registered in the same world coordinate frame used for robot poses.  
Specifically, we propose Radio-to-Pose Information (RPI),  an
interface that converts power residuals and their pose derivatives into a
Gaussian information matrix and vector pair. 
RPI builds on the insight that received power is sensitive to shared displacement in the world map.
The pose estimator uses the interface together with odometry and inter-robot measurements to update poses and minimize uncertainty.
We characterize the shared motions detectable from the map and observed links. For an uncertain shared map, RPI keeps correlations
across robots and time and reproduces joint Gaussian pose
estimates and covariances. 
Across $1000$ simulated Gaussian trials, RPI matches the results of joint estimation. 
Across 32 paired trials with a registered map, RPI reduces common-position RMSE from $0.542$\,m to $0.140$\,m in a $50\times50$\,m arena compared with a baseline that uses radio only to weight relative measurements. 

\end{abstract}

\section{Introduction}
To carry out tasks such as exploration, mapping, and environmental monitoring, robot teams rely on coordinated motion and shared observations. When global positioning is unavailable,
they can estimate motion through odometry and inter-robot measurements.
These observations constrain relative geometry, but the estimated team
can move away from its true location as a group. For example, translating every robot estimate by one meter leaves all pairwise distances unchanged.
This shared error matters when robots must reach mapped destinations or
report observations in world coordinates. 

It is also important to precisely estimate uncertainty because the team can know its relative geometry while remaining uncertain about its position in the world. This ambiguity is called \emph{gauge freedom}. 
%We determine \emph{gauge drift} for the temporal growth of error or uncertainty in these shared-motion directions. 
Gauge freedom describes indistinguishable coordinate frames, which in the presence of motion or noise leads to \emph{gauge drift}. Gauge drift describes 
the temporal growth of error or uncertainty in these shared-motion directions.
%whereas drift also depends on motion, noise, and the available reference measurements. 
Teams construct a reference frame from a prior over their initial poses, but subsequent noisy observations determine how well the team remains localized.
For trajectories without a fixed point of reference, common translations and rotations maintain relative observations, but they do not provide useful information about global state.
%A common translation and rotation of an unanchored trajectory preserves relative observations.
%An initial pose prior selects a frame, but subsequent observations determine how well the team remains localized within it.

\begin{figure}[t]
\centering
\includegraphics[width=\columnwidth]{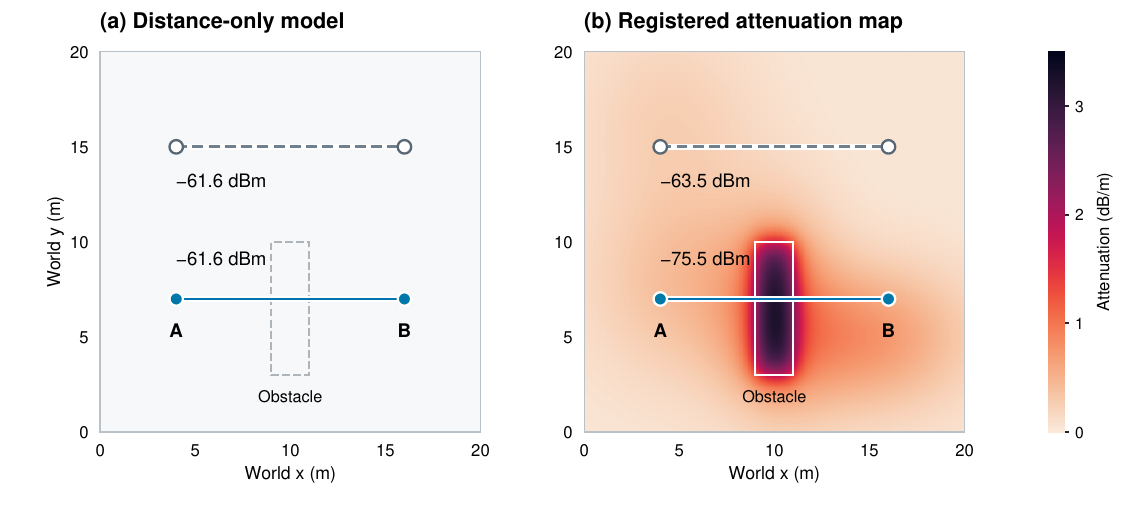}
\vspace{-20pt}
\caption{In the left map, when there is no registered attenuation map, equal-length links have identical distance-only predictions. In the right map, when a registered map is present, equal-length links have different power predictions. Colors indicate attenuation in dB/m.}
\label{fig:attenuation}
\end{figure}

A radio attenuation map aligned to the world frame can provide
a reference because received power depends on both link length
and the attenuation field. 
For example, in Fig.~\ref{fig:attenuation}, a common translation preserves separation and distance loss but changes link attenuation. Using a global attenuation map which captures knowledge of obstacles in the environment can help distinguish the team's positions by using differences in the received power they observe.

We propose the Radio-to-Pose Information (\name) interface which leverages this behavior.
The \name is an interface between two components: 
First, a radio module predicts received power from the current poses and attenuation map,
then determines the power mismatch and pose sensitivity as an information matrix and vector. 
Second, a pose estimator uses this message to compute a correction and its uncertainty. 

This paper describes three contributions:
\begin{enumerate}
\item We show that fixed weights on relative measurements add zero direct
information about common motion. We characterize which shared motions
the attenuation map and observed links can detect.
\item We propose the Radio-to-Pose Information (\name) interface, which converts power residuals and pose sensitivity into an information matrix and vector for pose estimation.
%We specify the correlations and message replacement needed to reproduce joint Gaussian pose estimates and covariances.
\item We derive centroid covariance growth and a lower bound on mean-squared common-position error under stated motion and measurement assumptions. We evaluate agreement with joint estimation,
drift correction, and sensitivity to map error.
\end{enumerate}
%The interface connects radio-map sensitivity to a pose update using Gaussian elimination and map-based anchoring.

\section{Related Work}
\noindent\textbf{Cooperative Localization.}
Relative measurements help robots estimate their positions with respect
to one another, but they do not determine the position and orientation of
the entire team in the world frame~\cite{roumeliotis2002,huang2011}. 
Gauge fixing chooses a coordinate frame for estimation~\cite{triggs2000}. Simultaneous
localization and mapping (SLAM) adds persistent environmental features and
loop closures, which can reduce trajectory drift~\cite{kaess2012}.
%SLAM can reduce trajectory drift by revisiting mapped features. 
However, a map built only from the robot's own measurements defines position relative to the robot's chosen coordinate frame rather than an externally registered world frame. 
Our comparison concerns a cooperative localizer that uses radio to set fixed weights on relative factors.

\noindent\textbf{Radio and Field Maps.}
Radio tomography estimates spatial attenuation from power measurements
and known link positions~\cite{wilson2010}. Radio SLAM jointly estimates
motion and propagation structure using radio observations, including
multipath information~\cite{leitinger2019,li2026}. 
Its reference frame depends on the available anchors and priors. Known physical anchors can provide a reference~\cite{li2026}, but may not be available. 
Instead, this paper focuses on studying
inter-robot links against a registered attenuation field. 
This follows the principle of localization against surveyed terrain or magnetic maps~\cite{carroll2021,ramos2022}.  % Adi: Removed repeated sentence. We determine how the observed links change based on the field and pass this information to the pose estimator.

\noindent\textbf{Gaussian Fusion.}
Factor-graph smoothers combine measurement models and priors through
local linearization~\cite{kaess2012}. 
Gaussian elimination provides a way to summarize part of this model for another estimator. 
Distributed smoothing must account for previously shared information when summaries are replaced~\cite{cunningham2013}. 
Fusion with shared variables must also
preserve the induced correlations or use an appropriate conservative
approximation~\cite{dagan2023,dagan2022}. 
\name applies these requirements
to a radio module whose uncertain map is shared across robots and time.
Equivalent radio factors in a factor-graph smoother give the same local
Gaussian result when the models and linearizations agree.

\section{Motion and Radio Model Setup}
\subsection{Shared Motion}
In the cooperative localization problem, a robot team seeks to estimate each robot's trajectory in 
%We seek to estimate the robot team’s trajectories in 
a fixed world frame over an estimation window. 
Consider $N$ robots over a finite set of time steps $\mathcal T=\{t_0,\ldots,t_w\}$, called the estimation window, where $t_0$ and $t_w$ are its first and last time steps.
Let $x_{i,t}=(p_{i,t},\theta_{i,t})\in\SE$ denote the true pose of robot $i$ at time $t\in\mathcal T$,
where $p_{i,t}\in\mathbb R^2$ is its position and
$\theta_{i,t}$ is its heading.
Let $X$ collect the poses of all robots across the estimation window, representing their trajectories. Each robot estimates its pose $\hat{x}_{i,t}$ using odometry and inter-robot measurements. A centralized server collects these estimates into $\hat{X}$ and solves for the team's trajectories. Both $X$ and $\hat{X}$ contain $N(w+1)$ poses.

In order to represent a gauge drift, a shared rigid transformation
$g=(R_\psi,t_g)\in\SE$ rotates every pose by $\psi$ and translates
every position by the translation vector $t_g\in\mathbb R^2$, where
$
R_\psi=
\begin{bmatrix}
\cos\psi&-\sin\psi\\
\sin\psi&\cos\psi
\end{bmatrix}.
$
\begin{figure}[t]
\centering
\includegraphics[width=\columnwidth]{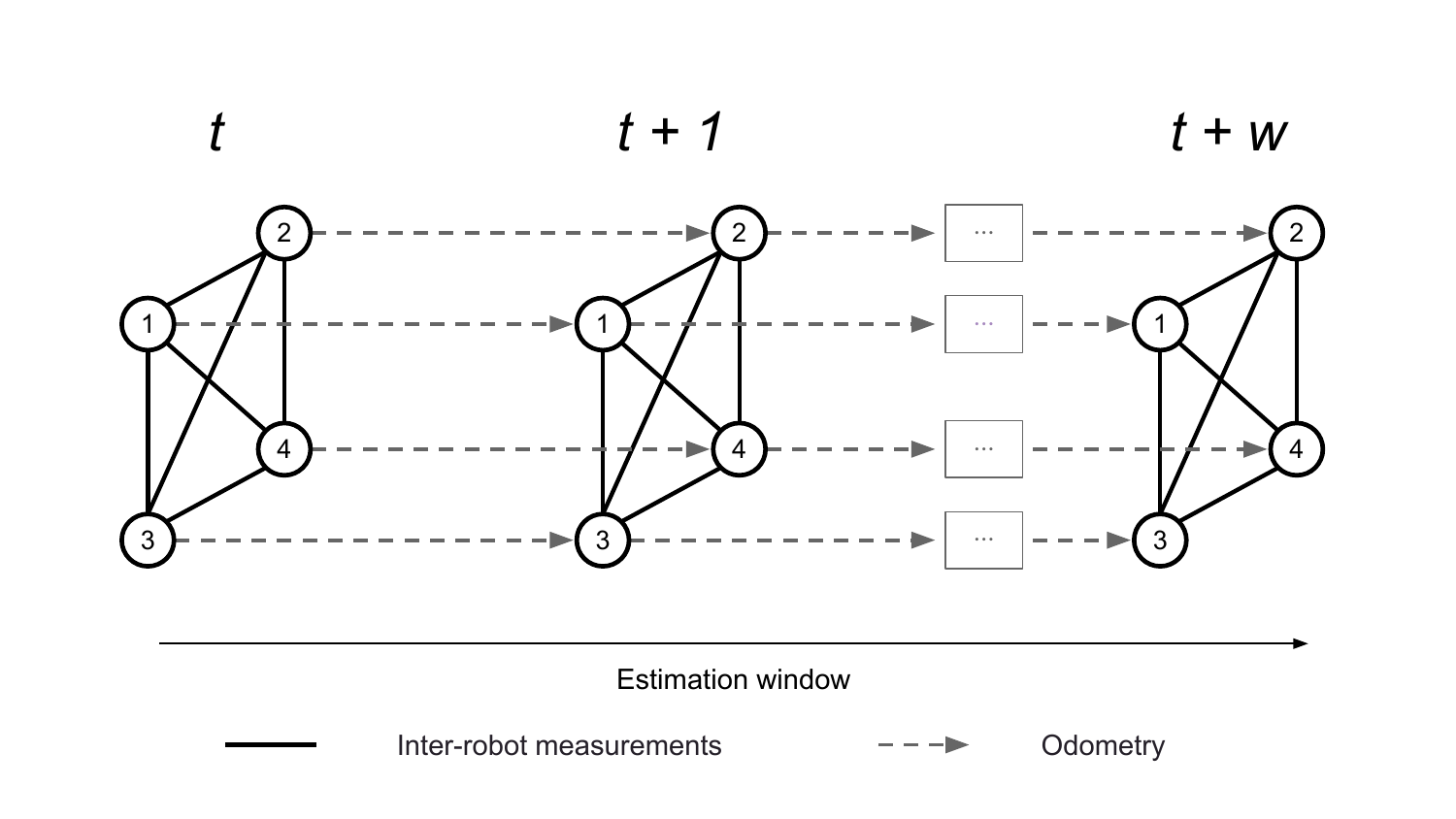}
\vspace{-30pt}
\caption{This graph represents the robot team at each time step. Solid links represent measurements between robots at the same time, while dashed arrows represent odometry of the same robot across consecutive time steps. Each residual $r_k(X)$ compares a recorded measurement with the measurement predicted by the poses.}
\label{fig:links}
\end{figure}
Thus, $g$ is applied to every pose in $X$,
giving $p_{i,t}\mapsto R_\psi p_{i,t}+t_g$ and
$\theta_{i,t}\mapsto\theta_{i,t}+\psi$.
This shared transformation preserves inter-robot distances, relative poses, and motion expressed in each robot's own frame. Let $\mathcal K$ denote the set of available relative measurements over the estimation window, including relative-position measurements between robots and odometry between consecutive time steps, as shown in Figure~\ref{fig:links}. Each index $k\in\mathcal K$ identifies one measurement and the number of available measurements is bounded by
$
|\mathcal K|
\le (w+1)\frac{N(N-1)}{2}+Nw
$.  For each $k$, let $r_k(X)$ denote the difference between the recorded sensor measurement and the corresponding value predicted using the robot poses in $X$. Since the transformation preserves these predicted values, $r_k(gX)=r_k(X)$. A shared translation changes the team's position in the fixed world frame, while a shared rotation changes its orientation. To determine which shared motions the measurements can detect,
we express small translations and rotations as pose perturbations.
About an origin $o$, with 90$^\circ$ rotation matrix
$S=\left[\begin{smallmatrix}0&-1\\1&0\end{smallmatrix}\right]$,
a small common translation $a$ and rotation $\omega$ produce
\begin{equation}
\delta p_{i,t}=a+\omega S(p_{i,t}-o),\quad
\delta\theta_{i,t}=\omega,\quad
\delta X=G\xi.
\label{eq:generators}
\end{equation}
where $\xi=(a_x,a_y,\omega)^\top$ describes the shared translation and rotation applied to all robots and time steps. The columns of $G\in\mathbb{R}^{3N(w+1)\times 3}$ correspond to the two translation directions and rotation.

\subsection{Radio Power and Reference}
For a radio link $\ell=(i,j,t)$ between robots $i$ and $j$ at time $t$,
let $d_{ij,t}=\|p_{j,t}-p_{i,t}\|_2>0$.
The field $F(v)$ provides attenuation in dB/m at world location
$v\in\mathbb R^2$. We predict received power by integrating the field over the spatial domain, weighted by each location’s contribution to the link.
$$
 \mu_\ell(X,F)=P_{\rm ref}-10\gamma\log_{10}(d_{ij,t}/d_0)
 -\int w_\ell(v,X)F(v)\,dv.
$$
$P_{\mathrm{ref}}$ is the reference received power in dBm
at distance $d_0=1\,\mathrm{m}$ without mapped attenuation,
and $\gamma$ is the path loss exponent governing the propagation decrease in received power with distance.
The weighting function $w_\ell(v,X)$ describes each location $v$'s contribution to link attenuation, given its position with respect to robot poses. We assume known propagation parameters and Gaussian power noise
$\epsilon\sim\mathcal N(0,R_{\mathrm{r}})$, where
$R_{\mathrm{r}}\succ0$, the radio measurement noise covariance matrix, is independent of the robot poses.
The measured power vector $y$, pose Jacobian $J_{\mathrm{r}}$,
and radio information matrix $M_{\mathrm{r}}$ satisfy
$$
y=\mu(X,F)+\epsilon,\quad
J_{\mathrm{r}}=\frac{\partial\mu}{\partial X},\quad
M_{\mathrm{r}}
=J_{\mathrm{r}}^\top R_{\mathrm{r}}^{-1}J_{\mathrm{r}},
$$
where $\epsilon\sim\mathcal N(0,R_{\mathrm{r}})$.
For a known field, the radio information along a pose
perturbation $u$ is
$u^\top M_{\mathrm{r}}u
=\|R_{\mathrm{r}}^{-1/2}J_{\mathrm{r}}u\|^2.
$
By registering the field $F$, we provide a reference in the world frame. If the weights satisfy $w_\ell(gv,gX)=w_\ell(v,X)$, transforming the poses to $X'=gX$ and the map to $F'(v)=F(g^{-1}v)$ leaves the predicted powers unchanged. Therefore, estimating the poses and map together cannot determine their absolute position and orientation unless the prior provides a world-frame reference. We provide this reference by setting the map’s position and orientation to the world frame. 

\section{Error and Map Information}
If gauge drift occurs, the robot team's collective position (i.e., the centroid) relative to the world map becomes uncertain
%Motion noise can cause error in the team's localization 
(\S\ref{s:common-pos-error-growth}).
We show that while existing techniques based on relative measurements alone cannot solve the gauge drift problem for robot teams (\S\ref{s:rel-weights}),
%and then 
%explain why cooperative localization via relative link weights does not address this source of error .
it is possible to use 
%information about the world map such as attenuation 
radio field attenuation information from the world map
to reduce this source of error 
%Finally, we show that world map information can reduce error from motion noise 
(\S\ref{s:map-sensitivity}).

%The Gaussian test propagates uncertainty in its attenuation values.
\subsection{Team Centroid Error Growth}
\label{s:common-pos-error-growth}
%With only relative observations, 
%Motion noise can cause uncertainty in the team centroid to grow over time. 
%We quantify this growth under the following motion model.
Relative observations do not measure the motion noise with respect to the centroid; 
as a result, the team cannot correct common-position drift using this kind of observation.
To see why, assume each robot's heading is known and its position evolves in world coordinates. Then, the position of robot $i$ at time step $t$ evolves to $t+1$ by $p_{i,t+1}=p_{i,t}+d_{i,t}+n_{i,t}$, where the process noise $n_{i,t}\sim\mathcal N(0,\sigma_p^2I_2)$,
$d_{i,t}$ is the displacement given by odometry per time step, and $\sigma_p^2$ is the motion-noise variance per position coordinate per time step.
In this equation, the process noises are independent across robots and time
and are independent of the initial state.

Define the team centroid as $c_t=N^{-1}\sum_{i=1}^{N}p_{i,t}$, where $N$ is the number of robots.
Assume that $c_0$ is the initial centroid and has
covariance $P_{c0}$. The measurement noise is independent of initial robot positions and process noise.

Averaging the robot dynamics gives
$
c_{t+1}
=
c_t+\bar d_t+\bar n_t,
\qquad
\bar n_t\sim
\mathcal N\left(0,\frac{\sigma_p^2}{N}I_2\right),
$
where $\bar d_t=N^{-1}\sum_i d_{i,t}$ and
$\bar n_t=N^{-1}\sum_i n_{i,t}$.
The average motion noise is independent of the motion noise that changes the robots’ relative positions.
The observation history does not reduce uncertainty in
$c_0$ or the accumulated centroid noise.
Thus,
$
\operatorname{Cov}(c_t\mid\mathcal Y_t)
=
P_{c0}+\frac{t\sigma_p^2}{N}I_2,
$
where $\mathcal Y_t$ denotes the observation history through time $t$.
For any estimate based on this history,
$
 \mathbb E\|\hat c_t-c_t\|^2
 \geq
 \operatorname{tr}P_{c0}
 +\frac{2t\sigma_p^2}{N}.
 \label{eq:lowerbound}
$

\subsection{Relative Weights}\label{s:rel-weights}
Relative observations improve estimates of the robots’ positions with respect to one another. However, using weighted links alone cannot solve this problem. 

If $r_k$ is invariant under the pose transformation $X\mapsto gX$, differentiating along a gauge direction gives $J_kG=0$, where $J_k=\partial r_k/\partial X$. For the joint pose–map transformation above, the corresponding identity is $J_{k,X}G_X+J_{k,F}G_F=0$. The weighted relative information matrix is
$M(q)=\sum_k q_kJ_k^\top R_k^{-1}J_k$, where $q_k$ describes nonnegative radio weights and $R_k$ is the indexed measurement noise covariance. Also, $G_X=G$ from Eq. \eqref{eq:generators} describes the shared pose
motion, and $G_F$ describes the corresponding map change caused by the same translation and rotation.
\begin{proposition}[Fixed relative weights]
\label{prop:rel_weights}
For fixed nonnegative radio weights $q_k$ on invariant relative measurements
with covariances $R_k\succ0$,
$$
G^\top M(q)G=\sum_kq_k(J_kG)^\top R_k^{-1}(J_kG)=0.
$$
Keeping only a subset of the relative factors still provides no information along the gauge directions.
\end{proposition}
\begin{proof}
Each term in the sum is zero, so keeping only some of the factors does not change the result.
\end{proof}
The proposition shows that fixed-weight relative measurements provide no direct information about shared translation or rotation. Additional pose information can come from modeling how received radio power depends on the robot poses.

\subsection{Map and Link Sensitivity}
\label{s:map-sensitivity}
A registered attenuation map can constrain this common motion when shared displacement changes the power predicted for the observed links.
Radio sensitivity to shared motion depends on how that motion
changes the attenuation sampled by the observed links.
Because a shared rigid motion leaves link length unchanged,
only the map attenuation term changes the predicted power.

Assume $F\in C^1$ and $w_\ell(gv,gX)=w_\ell(v,X)$. Then, substituting
$v=gv'$ gives
$
\int w_\ell(v,gX)F(v)\,dv
=
\int w_\ell(v',X)F(gv')\,dv'
$, where $v'=g^{-1}v$ is the spatial variable
before applying the rigid transformation $g$.
This expresses the attenuation using the original link weights
and the field evaluated at the transformed locations.
For $V_\xi(v')=a+\omega S(v'-o)$, differentiating along
the shared motion at the identity gives
\begin{equation}
 (J_{\mathrm r}G\xi)_\ell
 =-\int w_\ell(v',X)\nabla F(v')^\top V_\xi(v')\,dv'.
 \label{eq:derivative}
\end{equation}
$(J_{\mathrm r}G\xi)_\ell$ measures the sensitivity of link $\ell$'s predicted power to the shared motion $\xi$, with a fixed map. Let $\mathcal N_F$ denote the shared motions satisfying
$\nabla F(v')^\top V_\xi(v')=0$ for all $v'$.
The \emph{link sensitivity condition} requires every motion that the map reveals outside $\mathcal N_F$ to produce a nonzero first-order change in at least one observed link’s attenuation integral, despite possible missed variation or cancellation in other links.

\begin{proposition}[Detectable shared motion]
\label{prop:shared_motion}
Let $\mathcal L_{\mathrm{obs}}$ be the set of observed links.
With the map and the robots' relative geometry fixed,
$$
\mathcal N_F \subseteq
\left\{\xi : (J_{\mathrm r}G\xi)_\ell=0
\quad \forall\,\ell\in\mathcal L_{\mathrm{obs}}\right\}.
$$
The equality holds under the link sensitivity condition.
\end{proposition}
\begin{proof}
If $\xi\in\mathcal N_F$,
$\nabla F(v)^\top V_\xi(v)=0$ everywhere. By Eq.~\eqref{eq:derivative}, $(J_{\mathrm r}G\xi)_\ell=0$ for all
links, proving the inclusion. 
Under the link sensitivity condition, every 
$\xi\notin\mathcal N_F$ gives
$(J_{\mathrm r}G\xi)_\ell\ne0$ for at least one observed link,
establishing equality.
\end{proof}
With the map fixed and link weights satisfying
$w_\ell(gv,gX)=w_\ell(v,X)$, we use
Eq.~\eqref{eq:derivative} to construct several simple examples of unresolved motion. 
A constant field
has zero spatial gradient, so the radio model cannot detect any shared
translation or rotation. If $F(v)=f(v_x)$, vertical translation does
not change sampled link attenuation. 
A radial field cannot distinguish rotation about its center. 
Uncertain field values reduce the information carried by these derivatives. 
Repeated map patterns can produce the same radio measurements at different team locations. However, the full rank of the shared-motion Jacobian ensures local distinguishability. This means that any small shared translation or rotation of the team changes at least one predicted radio measurement.

\section{RPI Interface}
\label{sec:interface}
\begin{figure}[t]
\centering
\includegraphics[width=\columnwidth]{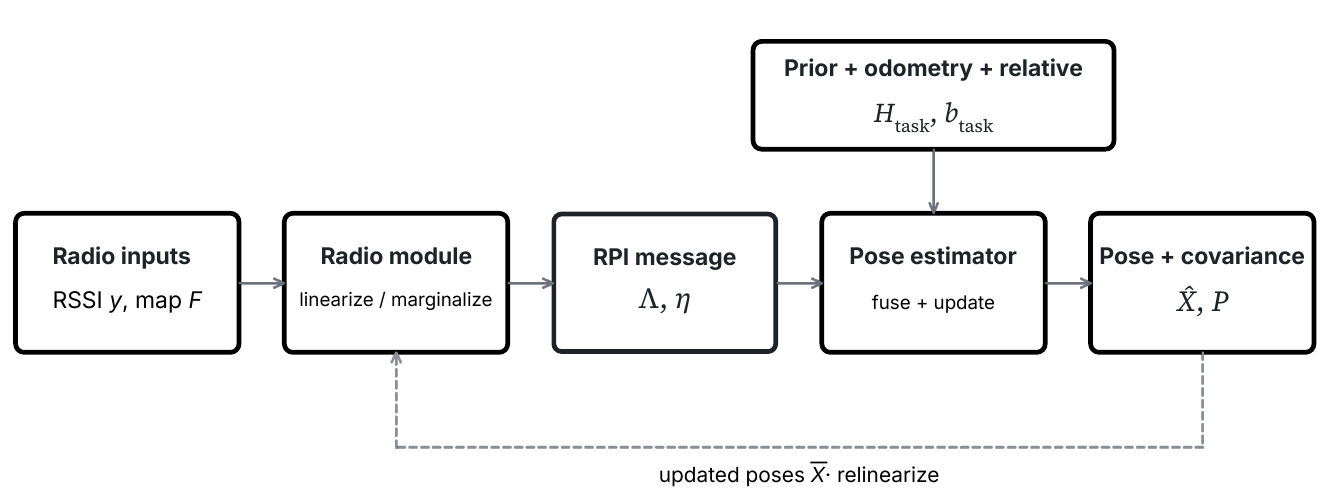}
\vspace{-25pt}
\caption{RPI converts power measurements and map information into
$(\Lambda,\eta)$ for pose estimation. Updated poses feed back to the radio module for relinearization. Revised messages replace previous contributions.}
\label{fig:rpi_flow}
\vspace{-15pt}
\end{figure}

We design \name using the insight that a radio map aligned to world coordinates can provide a reference for the team’s position. We treat the map’s position and orientation in world coordinates as known. 
This method accounts for Gaussian uncertainty in its attenuation values as we describe in \S\ref{s:power-to-pose}.
We introduce \name to pass information from the radio module to the pose estimator, as Fig.~\ref{fig:rpi_flow} illustrates. 
%Fig.~\ref{fig:rpi_flow} shows how the radio module forms an information matrix and vector, and the pose estimator uses updated poses to relinearize.

\subsection{Grid model}

We first discretize the attenuation field and compute received power over it.
To support nonlinear settings, where the predicted radio power depends nonlinearly on robot positions, we represent the attenuation field by grid values $\mathbf F$.
Within each cell, bilinear interpolation combines its four corner values with weights $\beta_m(v)$ that sum to one, giving
$F_h(v)=\sum_m\beta_m(v)F_m$, where $h$ denotes the grid spacing and $F_m$ the attenuation value at grid vertex $m$. We then sample this field along each
radio link to obtain $A_h(X)$, the matrix that converts map grid values into attenuation for each link $\ell=(i,j,t)$. Attenuation and distance loss use
$\widetilde d_{ij,t}
=\sqrt{\|p_{j,t}-p_{i,t}\|^2+\delta_d^2}$,
where $\delta_d\approx10^{-6}$\,m.
$$
\begin{aligned}
[A_h(X)]_{\ell m}
&= \sum_{q=1}^{Q}\alpha_{\ell q}(X)
   \beta_m(v_{\ell q}(X)),\\
[\mu_{\rm free}(X)]_\ell
&= P_{\rm ref}
   -10\gamma\log_{10}\!\left(
   \frac{\widetilde d_{ij,t}}{d_0}\right),\\
\mu_h(X,\mathbf F)
&= \mu_{\rm free}(X)-A_h(X)\mathbf F.
\end{aligned}
$$
The vector $\mu_{\rm free}(X)$ contains the predicted received
powers before mapped attenuation, with entry
$[\mu_{\rm free}(X)]_\ell$ for each link
$\ell\in\mathcal L_{\rm obs}$. $\alpha_{\ell q}$ is the integration weight of sample $q$ along link $\ell$. 
We use $Q=64$ samples including both endpoints,
$v_{\ell q}=(1-s_q)p_{i,t}+s_qp_{j,t}$, where $s_q=\frac{q-1}{Q-1}$,
with weights
$
\alpha_{\ell q}
=\widetilde d_{ij,t}\frac{b_q}{\sum_s b_s},\qquad
b_q=\frac{1}{1+e^{-\kappa[1-(2s_q-1)^2]}}$. $\kappa$ is the spatial weight rise parameter that dictates how fast the weight gradient changes between the mid and endpoints. We set $\kappa=8$ to give nearly uniform weights across the link interior while reducing the endpoint weights.
We differentiate the discretized power prediction with respect to the poses at the current estimates, holding the grid values fixed, to obtain the radio pose Jacobian $J_X$. Then, we differentiate the prediction with respect to the grid values holding the poses fixed, which gives $J_F=-A_h(X)$.

\subsection{Power to Pose Information}\label{s:power-to-pose}

We show how the radio module converts received power into an informative pose update.
Let $\mathbf F$ denote the map values and $\hat{\mathbf F}$ their current estimate.
At the current pose and map estimates $(\hat X,\hat{\mathbf F})$, define the power residual
$r=y-\mu_h(\hat X,\hat{\mathbf F})$.
 Assume a map prior
$\mathbf F\sim\mathcal N(\bar{\mathbf F},P_F)$,
independent of the pose prior, where $\bar{\mathbf F}$ is
the prior map mean and $P_F\succ0$ its covariance.
Let $z$ denote a local pose correction and
$f=\mathbf F-\hat{\mathbf F}$ a map correction.
Linearization gives $r\approx J_Xz+J_Ff+\epsilon$.
Then $f\sim\mathcal N(m_f,P_F)$, where $m_f=\bar{\mathbf F}-\hat{\mathbf F}$. 
Let $H_{\rm task}$ denote the pose information matrix and
$b_{\rm task}$ the information vector driving the latest pose correction, both formed from the linearized odometry, relative factors, and pose priors. {$\rm task$} refers
to the pose-estimation objective before adding radio information.
Joint Gaussian estimation solves,
with $W=R_r^{-1}$ where $W$ is the precision matrix that describes how strongly radio prediction errors are penalized in the estimator,
\begin{equation}
\begin{aligned}
\min_{z,f}\quad&
\tfrac12 z^\top H_{\rm task}z-b_{\rm task}^\top z\\
&+\tfrac12\|r-J_Xz-J_Ff\|_W^2
+\tfrac12\|f-m_f\|_{P_F^{-1}}^2,
\end{aligned}
\label{eq:jointobjective}
\end{equation}
The radio residual is weighted by $W$, and the map-prior
deviation by the precision matrix $P_F^{-1}$, the inverse
of the map-prior covariance.
To pass the radio information
to a pose-only estimator, define
$$
\begin{aligned}
A&=J_X^\top WJ_X,
& a&=J_X^\top Wr,\\
B&=J_F^\top WJ_F+P_F^{-1},
& b&=J_F^\top Wr+P_F^{-1}m_f,\\
C&=J_X^\top WJ_F.
\end{aligned}
$$
Eliminating the map correction $f$ through the Schur
complement of the map information $B$ gives the
RPI pair

\begin{equation}
\Lambda=A-CB^{-1}C^\top,\qquad
\eta=a-CB^{-1}b.
\label{eq:schur}
\end{equation}
With a known map, we set $\hat{\mathbf F}$ to its known value, so $f=0$ and $(\Lambda,\eta)=(A,a)$.
Adding task information gives
the Gaussian pose covariance and update
\begin{equation}
P_z=(H_{\rm task}+\Lambda)^{-1},\qquad
\hat z=P_z(b_{\rm task}+\eta).
\label{eq:poseupdate}
\end{equation}
The summed information must be positive definite. For nonlinear iterations, the estimator solves for
$\hat z$, updates the poses, and relinearizes.

\begin{proposition}[Gaussian pose agreement]
\label{prop:pose_agree}
With matching factors, priors, weights, coordinates, and
linearizations, Eq.~\eqref{eq:schur}--\eqref{eq:poseupdate}
reproduce the pose mean and covariance obtained by solving
Eq.~\eqref{eq:jointobjective} jointly.
\end{proposition}
\begin{proof}
The joint objective is quadratic in $(z,f)$. Completing the
square in $f$ and integrating it leaves
$\tfrac12z^\top(H_{\rm task}+\Lambda)z
-(b_{\rm task}+\eta)^\top z$ plus a constant. The minimizer
and covariance are given by Eq.~\eqref{eq:poseupdate}.
\end{proof}

\subsection{Shared History}
Uncertainty in the shared map can affect pose estimates for multiple robots at different times. \name therefore combines the radio measurements across robots and time steps before eliminating the map variables. This maintains the cross-robot and cross-time correlations the shared map induces. 
With respect to Proposition~\ref{prop:pose_agree}, the resulting message gives the exact pose marginal for the Gaussian system.

The pose estimator reconstructs the \name interface after each relinearization. 
For $N$ robots and $L$ time steps, a pose information matrix requires $O(N^2L^2)$ storage. The \name message contains $(\Lambda,\eta)$, the order of the pose variables, and the linearization point. 

\begin{figure}[t]
\centering
\includegraphics[width=\columnwidth]{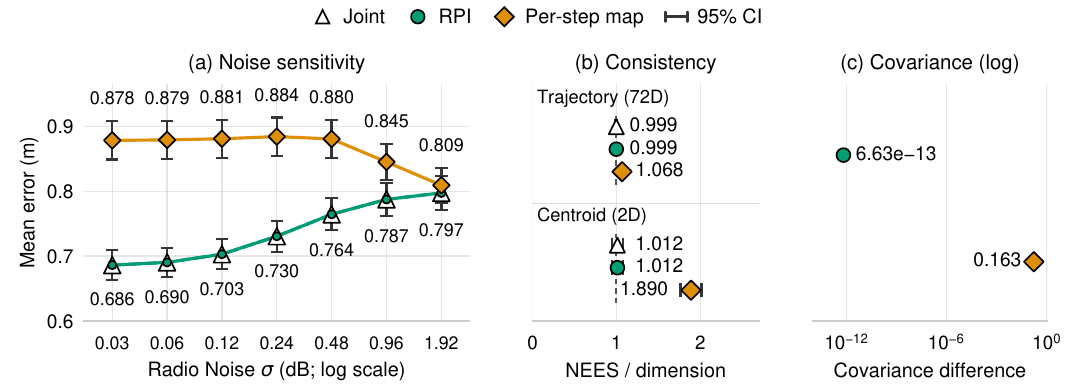}
\caption{Gaussian shared-map test over 1,000 paired trials.
(a) Mean terminal centroid error versus radio-noise standard deviation.
(b) Trajectory and terminal-centroid NEES per coordinate,
with nominal value one.
(c) Covariance difference relative to Joint.
Error bars are 95\% intervals from trial means.}
\label{fig:gaussian}
\end{figure}

\section{Evaluation}

We evaluate how closely RPI matches joint Gaussian estimation (\S\ref{s:linear}), its ability to correct nonlinear drift
(\S\ref{s:nonlinear}), its sensitivity to map error
(\S\ref{s:map-error}), and the directions constrained by the map
(\S\ref{s:ablation}).

\subsection{Methodology}

All evaluations generate noisy measurements from known robot
trajectories and an attenuation map, then estimate positions
with known headings. 
Methods within each trial share observations and initialization.

For position error $e_{i,t}=\hat p_{i,t}-p_{i,t}$, we report
common error $\bar e_t=N^{-1}\sum_i e_{i,t}$ and relative
error $e_{i,t}-\bar e_t$, measuring centroid displacement
and errors in the team’s relative arrangement, respectively.
We evaluate nonlinear RMSE over the final third of each trajectory to emphasize accumulated drift later in the run.

For stacked terminal-position covariance $P$, common
uncertainty is $\sqrt{\operatorname{tr}P_{\mathrm{centroid}}}$,
where
$
P_{\mathrm{centroid}}=C_pPC_p^\top$, $
C_p=N^{-1}(\mathbf1_N^\top\otimes I_2).$
Centroid NEES per coordinate is
$\bar e_{t_w}^\top P_{\mathrm{centroid}}^{-1}\bar e_{t_w}/2$,
with expected value one under a calibrated Gaussian model. Larger values indicate overconfidence. We also evaluate
coverage of 95\% coordinate intervals in the Gaussian
test.

\subsection{Linear Gaussian Interface Test}\label{s:linear}

We test whether RPI preserves the joint Gaussian pose solution and whether shared-map correlations affect accuracy.
Across 1,000 trials, we record three robots at 12 time steps, giving 72 position coordinates. 
The measurement Jacobians remain fixed. For this test, we use a simplified map with nine uncertain radial-basis coefficients governed by a spatially correlated Gaussian prior. We set radio, relative-position, and map-coefficient standard deviations to be $0.12$\,dB, $0.06$\,m, and $0.25$\,dB/m, respectively. We vary radio standard deviation from 0.03 to 1.92 dB in the noise-sensitivity experiment.

We consider three configurations:
\begin{inparaenum}[(i)]
\item \emph{Joint} estimates robot positions and one shared map together. 
\item \emph{RPI} eliminates the map variables and passes their contribution to the pose solver through $(\Lambda,\eta)$.
\item \emph{Per-step map} assigns an independent map to each time step, preserving within-step correlations but discarding correlations across time. 
\end{inparaenum}

We expect RPI to outperform Per-step map because it preserves shared-map correlations across time. Similarly, if RPI is a useful abstraction, it would match Joint's performance.
Fig.~\ref{fig:gaussian} confirms these expectations.
RPI and Joint overlap across radio-noise standard deviations from $0.03$ to $1.92$\,dB, with mean terminal centroid error increasing from $0.686$ to $0.797$\,m. 
As the noise level increases, all three configurations converge to the same pose priors and relative-position measurements.

%Per-step map has higher error throughout, but its error decreases at higher noise levels, narrowing the gap to RPI and Joint. 
%This behavior is consistent with noisier radio measurements receiving less weight, reducing the effect of incorrectly modeled shared-map correlations. 

Per-step map also underestimates uncertainty. 
Terminal centroid NEES is $1.890$, compared with $1.012$ for RPI and Joint, and nominal 95\% coordinate coverage falls from approximately $94.7\%$ to $84.88\%$.
%Overall, RPI captures the information of Joint
%Thus, RPI preserves the joint solution, while discarding shared-map history reduces accuracy and produces overconfidence.

\subsection{Nonlinear Drift Correction}\label{s:nonlinear}

We test whether RPI reduces common drift
while preserving relative accuracy. We simulate four robots moving for
80 steps of $0.5$\,s and observe all six inter-robot links.
Known headings and world-coordinate relative-position measurements
leave two common translation directions unconstrained by the
relative measurements.
We update radio predictions and Jacobians as the estimates change.
Methods share relative-factor weights within each map condition,
and these weights remain unchanged during optimization.

We compare four methods using the same supplied map,
observations, initialization, and fixed relative-factor weights:
\begin{inparaenum}[(i)]
\item \emph{Scalar} uses radio quality only to reweight
relative-position factors.
\item \emph{RPI (fixed map)} treats the supplied map as exact
and adds pose information $(\Lambda,\eta)=(A,a)$.
\item \emph{Joint (fixed map)} directly adds the nonlinear
radio term
$\tfrac12\|y-\mu_h(X,\mathbf F)\|_{R_{\mathrm r}^{-1}}^2$
and optimizes the poses with $\mathbf F$ fixed.
\item \emph{RPI (uncertain map)} uses the shared
attenuation-value model in Section~\ref{sec:interface}.
\end{inparaenum}

The uncertain-map method estimates all $40\times40$ grid
values with a Gaussian prior centered on the supplied map.
We use a coefficient standard deviation of $0.25$\,dB/m
and Gaussian spatial correlations with a $3$\,m length scale.
The variance combines an 80\% spatially correlated component
and a 20\% component independent across grid vertices.
This covariance remains unchanged across map conditions.
We use up to 25 Levenberg--Marquardt iterations, obtaining
map increments after each pose update before relinearizing.
Terminal uncertainty uses the terminal-position block of
the undamped local marginal pose covariance $P_z$ in
Eq.~\eqref{eq:poseupdate}.

Across 32 paired trials, initial-position, odometry,
relative-coordinate, and radio noise standard deviations
are $0.5$\,m, $0.08$\,m, $0.1$\,m, and $3$\,dB.
To standardize the test, all methods share fixed radio quality weights on relative factors. 

\begin{figure}[t]
\centering
\includegraphics[width=\columnwidth]{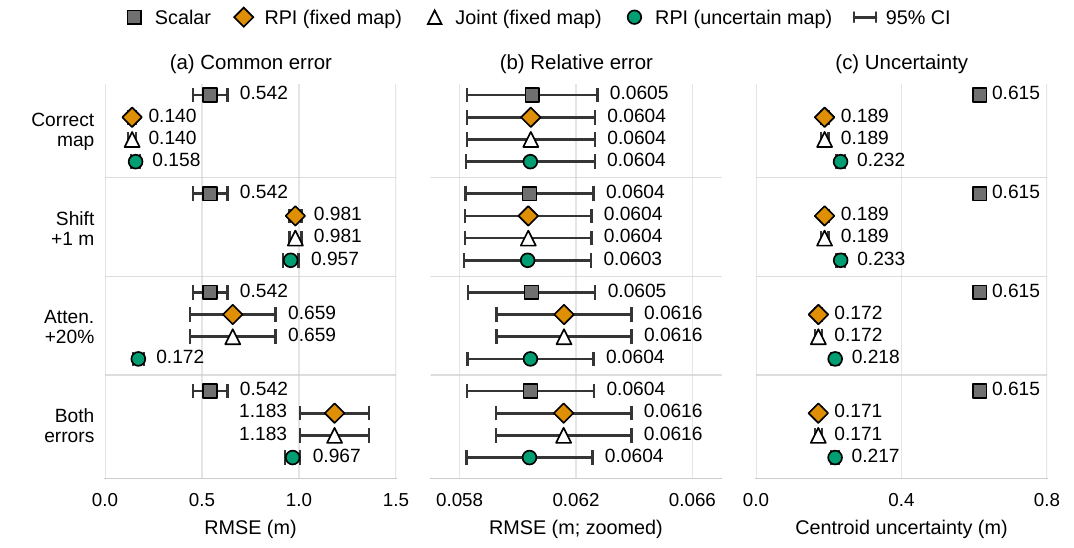}
\caption{Results from 32 paired nonlinear trials with correct
and perturbed maps: (a) common-position RMSE,
(b) relative-position RMSE, and (c) terminal common uncertainty.
RMSE uses the final third of each trajectory.
Error bars show 95\% intervals computed from trial means.}
\label{fig:drift}
\end{figure}

We expect that fixed-map RPI will again match Joint's
common error while preserving its relative accuracy,
and will reduce common-position RMSE relative to Scalar.
Fig.~\ref{fig:drift} confirms these results:
fixed-map RPI and Joint reduce common-position RMSE
from Scalar's $0.542$\,m to $0.140$\,m, a $74\%$ reduction,
with paired-change interval $[-0.494,-0.311]$\,m.
Relative RMSE remains near $0.060$\,m.
For fixed-map RPI and Joint, terminal common uncertainty
decreases from Scalar's $0.615$\,m to $0.189$\,m.
Uncertain-map RPI gives common-position RMSE of
$0.158$\,m and terminal common uncertainty of $0.232$\,m.

\subsection{Map Error Ablation}\label{s:map-error}

We test sensitivity to map error by keeping the
true map unchanged while supplying a map shifted by
$1$\,m, attenuation values multiplied by $1.2$, or both.
Relative-factor weights are recomputed for each supplied
map and shared across methods. Fixed-map RPI and Joint
treat the supplied map as exact, while uncertain-map RPI
uses the same attenuation-value model and prior covariance
as in \S\ref{s:nonlinear}.

Fig.~\ref{fig:drift} shows that fixed-map RPI and Joint
overlap, with final-third common-position RMSE of
$0.981$, $0.659$, and $1.183$\,m under registration,
attenuation, and combined errors, respectively.
Uncertain-map RPI reduces these errors to $0.957$,
$0.172$, and $0.967$\,m, while maintaining relative
RMSE near $0.060$\,m, and outperforms Scalar's
$0.542$\,m under attenuation error, but not under
registration or combined errors. This occurs because uncertain-map RPI models attenuation-value
uncertainty but does not explicitly model registration error, which can bias the position estimates.

Fixed-map RPI and Joint report terminal centroid
uncertainty between $0.171$ and $0.189$\,m despite
their increased position errors. Uncertain-map RPI
reports between $0.217$ and $0.233$\,m.
Thus, shared attenuation uncertainty improves robustness
to attenuation-scale error but does not resolve
registration bias in these experiments.
\subsection{Directional Map Ablation}\label{s:ablation}

Finally, we test which translation directions the map can constrain
using constant, $x$-varying, and two-dimensional maps.
At the supplied map means, we verify that the observed links provide information about zero, one, and two independent directions of shared translation, respectively. Trajectories, non-radio measurements, initialization, and radio-noise realizations remain fixed.
Each map generates its own radio observations
and is also used for inference, avoiding map mismatch.
We compare Scalar, fixed-map RPI, and uncertain-map RPI
using the nonlinear procedure in \S\ref{s:nonlinear}.

\begin{figure}[t]
\centering
\includegraphics[width=\columnwidth]{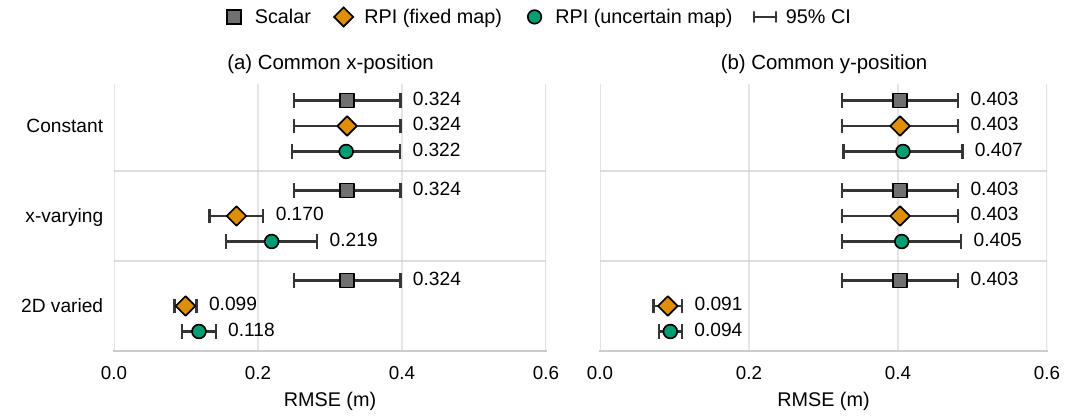}
\caption{Directional map ablation over 32 paired trials.
Final-third common $x/y$ RMSE for maps with
translation-information ranks zero, one, and two.
Error bars show 95\% intervals across trials.}
\label{fig:directional}
\end{figure}

Fig.~\ref{fig:directional} shows that Scalar has common
$x/y$ RMSE of $0.324/0.403$\,m in all conditions.
Fixed-map RPI gives the same errors with the constant map,
$0.170/0.403$\,m with the $x$-varying map, and
$0.099/0.091$\,m with the two-dimensional map.
Uncertain-map RPI gives $0.322/0.407$, $0.219/0.405$,
and $0.118/0.094$\,m, respectively, preserving the
directional pattern with slightly higher mean errors
in the useful directions.
Both RPI variants therefore primarily reduce error
in directions informed by the map.

\section{Conclusion}
In this work, we showed how a registered attenuation map
provides information about shared robot motion that fixed
weights on relative measurements cannot directly supply.
We introduced \name, a mathematical interface that converts radio measurements and map information into pose information while preserving shared-map correlations. Our analysis characterizes which shared motions the map and observed links can detect.

Our simulations confirm that \name matches the pose estimates and covariances from the benchmarked joint optimization method. Our nonlinear experiments
show reduced common-position error while maintaining
relative-position accuracy, with directional map ablations confirming that correction occurs in map-constrained directions. Map mismatch can increase position error without a corresponding increase in reported uncertainty. Future work will evaluate \name on real measured radio data and study pose-update convergence. We will release the code upon acceptance.

\bibliographystyle{ieeetr}
\bibliography{references}

@article{roumeliotis2002,
  author={Roumeliotis, S.I. and Bekey, G.A.},
  journal={IEEE Transactions on Robotics and Automation}, 
  title={Distributed multirobot localization}, 
  year={2002},
  volume={18},
  number={5},
  pages={781-795},
  doi={10.1109/TRA.2002.803461}}

@article{huang2011,
  title    = "Observability-based consistent {EKF} estimators for multi-robot
              cooperative localization",
  author   = "Huang, Guoquan P and Trawny, Nikolas and Mourikis, Anastasios I
              and Roumeliotis, Stergios I",
  journal  = "Autonomous Robots",
  volume   =  30,
  number   =  1,
  pages    = "99--122",
  month    =  jan,
  year     =  2011
}

@InProceedings{triggs2000,
author="Triggs, Bill
and McLauchlan, Philip F.
and Hartley, Richard I.
and Fitzgibbon, Andrew W.",
editor="Triggs, Bill
and Zisserman, Andrew
and Szeliski, Richard",
title="Bundle Adjustment --- A Modern Synthesis",
booktitle="Vision Algorithms: Theory and Practice",
year="2000",
publisher="Springer Berlin Heidelberg",
address="Berlin, Heidelberg",
pages="298--372",
isbn="978-3-540-44480-0"
}

@article{wilson2010, author = {Wilson, Joey and Patwari, Neal}, title = {Radio Tomographic Imaging with Wireless Networks}, year = {2010}, issue_date = {May 2010}, publisher = {IEEE Educational Activities Department}, address = {USA}, volume = {9}, number = {5}, issn = {1536-1233}, url = {https://doi.org/10.1109/TMC.2009.174}, doi = {10.1109/TMC.2009.174}, journal = {IEEE Transactions on Mobile Computing}, month = may, pages = {621–632}, numpages = {12} }

@article{leitinger2019,
   title={A Belief Propagation Algorithm for Multipath-Based SLAM},
   volume={18},
   ISSN={1558-2248},
   url={http://dx.doi.org/10.1109/TWC.2019.2937781},
   DOI={10.1109/twc.2019.2937781},
   number={12},
   journal={IEEE Transactions on Wireless Communications},
   publisher={Institute of Electrical and Electronics Engineers (IEEE)},
   author={Leitinger, Erik and Meyer, Florian and Hlawatsch, Franz and Witrisal, Klaus and Tufvesson, Fredrik and Win, Moe Z.},
   year={2019},
   month=Dec, pages={5613–5629} }

@misc{li2026,
      title={Probabilistic Occupancy Grid for Radio-Based SLAM}, 
      author={Xuhong Li and Erik Leitinger and Fredrik Tufvesson and Florian Meyer},
      year={2026},
      eprint={2603.03559},
      archivePrefix={arXiv},
      primaryClass={eess.SP},
      url={https://arxiv.org/abs/2603.03559}, 
}

@article{kaess2012,
   author = {M. Kaess and H. Johannsson and R. Roberts and V. Ila and J.J.
	Leonard and F. Dellaert},
   title = {{iSAM2}: Incremental Smoothing and Mapping Using the {B}ayes Tree},
   journal = {Intl. J. of Robotics Research, IJRR},
   volume = {31},
   number = {2},
   pages = {216-235},
   month = feb,
   year = {2012}
}

@INPROCEEDINGS{cunningham2013,
  author={Cunningham, Alexander and Indelman, Vadim and Dellaert, Frank},
  booktitle={2013 IEEE International Conference on Robotics and Automation}, 
  title={DDF-SAM 2.0: Consistent distributed smoothing and mapping}, 
  year={2013},
  volume={},
  number={},
  pages={5220-5227},
  doi={10.1109/ICRA.2013.6631323}}

@article{dagan2023,
author = {Dagan, Ofer and Ahmed, Nisar R.},
title = {Exact and Approximate Heterogeneous Bayesian Decentralized Data Fusion},
year = {2023},
issue_date = {April 2023},
publisher = {IEEE Press},
volume = {39},
number = {2},
issn = {1552-3098},
url = {https://doi.org/10.1109/TRO.2022.3226115},
doi = {10.1109/TRO.2022.3226115},
journal = {Trans. Rob.},
month = apr,
pages = {1136–1150},
numpages = {15}
}

@INPROCEEDINGS{dagan2022,
  author={Dagan, Ofer and Ahmed, Nisar R.},
  booktitle={2022 IEEE/RSJ International Conference on Intelligent Robots and Systems (IROS)}, 
  title={Conservative Filtering for Heterogeneous Decentralized Data Fusion in Dynamic Robotic Systems}, 
  year={2022},
  volume={},
  number={},
  pages={5840-5847},
  doi={10.1109/IROS47612.2022.9981414}}

@article {carroll2021,
author = {Carroll, Jason D. and Canciani, Aaron J.},
title = {Terrain-referenced navigation using a steerable-laser measurement sensor},
volume = {68},
number = {1},
pages = {115--134},
year = {2021},
doi = {10.1002/navi.406},
publisher = {Institute of Navigation},
issn = {0028-1522},
URL = {https://navi.ion.org/content/68/1/115},
eprint = {https://navi.ion.org/content/68/1/115.full.pdf},
journal = {NAVIGATION: Journal of the Institute of Navigation}
}

@INPROCEEDINGS{ramos2022,
  author={Ramos, J. Humberto and Shin, Jaejeong and Volle, Kyle and Buzaud, Paul and Brink, Kevin and Ganesh, Prashant},
  booktitle={2022 IEEE/RSJ International Conference on Intelligent Robots and Systems (IROS)}, 
  title={Information-Aware Guidance for Magnetic Anomaly based Navigation}, 
  year={2022},
  volume={},
  number={},
  pages={6347-6354},
  doi={10.1109/IROS47612.2022.9981709}}
\end{document}